\documentclass[10pt, conference, compsocconf]{IEEEtran}
\IEEEoverridecommandlockouts
\usepackage{amsmath}
\usepackage{algorithmic}
\usepackage[linesnumbered]{algorithm2e}
\usepackage{color}
\usepackage{multirow}
\usepackage{booktabs}
\usepackage[hyphens]{url}
\usepackage{breakurl}

\usepackage[usenames,dvipsnames]{xcolor}

\usepackage[bookmarks=false]{hyperref}
\hypersetup{citecolor=blue,linkcolor=blue}

\usepackage{graphicx}
\usepackage{listings}
\usepackage{setspace}
\usepackage{caption}
\usepackage{comment}
\usepackage{floatrow}
\usepackage{subfigure}
\usepackage[most]{tcolorbox}

\usepackage{endnotes,microtype,xspace,graphicx,fancyvrb,multirow}
\usepackage{fancyhdr}
\usepackage[normalem]{ulem}
\usepackage[hyphens]{url}

\usepackage{array}
\newcommand{\subparagraph}{}
\usepackage[compact]{titlesec}
\titleformat{\paragraph}[runin]{\vspace{-3pt}\normalfont\normalsize\bfseries\small\sffamily
  }{}
  {0pt}{}[.]
  
\titleformat{\subparagraph}[runin]{\normalfont\normalsize\itshape
 }{}
 {0pt}{}{}

\newcommand{\TK}[1]{\textcolor{blue}{TK: #1}}
\newcommand{\Parth}[1]{\textcolor{brown}{Parth: #1}}

\newcommand{\insertFigure}[2]{
    \begin{figure}[t]
\setlength{\abovecaptionskip}{0pt}
\setlength{\belowcaptionskip}{0pt}
        \centering
        \includegraphics[width=\linewidth]{figs/#1.pdf}
	\vspace{-2mm}
        \caption{\small #2}
	\vspace{-2mm}
        \label{fig:#1}
    \end{figure}
}

\newcommand{\insertPNG}[2]{
    \begin{figure}[t]
\setlength{\abovecaptionskip}{0pt}
\setlength{\belowcaptionskip}{0pt}
        \centering
        \includegraphics[width=\linewidth]{figs/#1.PNG}
        \caption{#2}
        \label{fig:#1}
    \end{figure}
}

\newcommand{\insertWideFigure}[2]{

    \begin{figure*}[h]
\setlength{\abovecaptionskip}{0pt}
\setlength{\belowcaptionskip}{0pt}
        \centering
        \includegraphics[width=\textwidth]{figs/#1.pdf}
	\vspace{-4mm}
        \caption{\small #2}
	\vspace{-2mm}
        \label{fig:#1}
    \end{figure*}

}

\newcommand{\insertWidePNG}[2]{
    \begin{figure*}[h]
\setlength{\abovecaptionskip}{0pt}
\setlength{\belowcaptionskip}{0pt}
        \centering
        \includegraphics[width=\textwidth]{figs/#1.PNG}
	\vspace{-4mm}
        \caption{\small #2}
	\vspace{-2mm}
        \label{fig:#1}
    \end{figure*}
}

\newcommand{\squishlist}{
 \begin{list}{$\bullet$}
  { \setlength{\itemsep}{0pt}
     \setlength{\parsep}{3pt}
     \setlength{\topsep}{3pt}
     \setlength{\partopsep}{0pt}
     \setlength{\leftmargin}{1.5em}
     \setlength{\labelwidth}{1em}
     \setlength{\labelsep}{0.5em} } }

\newcommand{\squishlisttwo}{
 \begin{list}{$\bullet$}
  { \setlength{\itemsep}{0pt}
     \setlength{\parsep}{0pt}
    \setlength{\topsep}{0pt}
    \setlength{\partopsep}{0pt}
    \setlength{\leftmargin}{2em}
    \setlength{\labelwidth}{1.5em}
    \setlength{\labelsep}{0.5em} } }

\newcommand{\squishend}{
  \end{list}  }

\usepackage{cite}
\usepackage{amsmath,amssymb,amsfonts}
\usepackage{algorithmic}
\usepackage{graphicx}
\usepackage{textcomp}
\usepackage{xcolor}
\usepackage{comment}
\def\BibTeX{{\rm B\kern-.05em{\sc i\kern-.025em b}\kern-.08em
    T\kern-.1667em\lower.7ex\hbox{E}\kern-.125emX}}
    
\begin{document}


\title{CLAN: Continuous Learning using Asynchronous Neuroevolution on Commodity Edge Devices
\thanks{This work was performed while Parth Mannan was a graduate student at Georgia Institute of Technology.}}
\linespread{0.95}
\author{\IEEEauthorblockN{Parth Mannan}
\IEEEauthorblockA{NVIDIA\\
Santa Clara, CA\\
parth.mannan95@gmail.com}
\and
\IEEEauthorblockN{Ananda Samajdar}
\IEEEauthorblockA{Georgia Institute of Technology\\
Atlanta, GA\\
anandsamajdar@gatech.edu}
\and
\IEEEauthorblockN{Tushar Krishna}
\IEEEauthorblockA{Georgia Institute of Technology\\
Atlanta, GA\\
tushar@ece.gatech.edu}
}

\maketitle
\begin{abstract}
Recent advancements in machine learning algorithms, especially the development of Deep Neural Networks (DNNs) have transformed the landscape of Artificial Intelligence (AI). 
With every passing day, deep learning based methods are applied to solve new problems with exceptional results.

The portal to the real world is the edge.
The true impact of AI 
can only be fully realized if we can have AI agents continuously interacting with the real world and solving everyday problems.
Unfortunately, high compute and memory requirements of DNNs acts a huge barrier towards this vision.
Today we circumvent this problem by deploying special purpose inference hardware on the edge while procuring trained models from the cloud.
This approach, however, 
relies on constant interaction with the cloud for transmitting all the data, 
training on massive GPU clusters,
and downloading updated models. This is challenging 
for bandwidth, privacy, 
and constant connectivity concerns that autonomous agents may exhibit.

In this paper we evaluate techniques for enabling  adaptive intelligence on edge devices with zero interaction with any high-end cloud/server.
We build a prototype distributed system of Raspberry Pis communicating via WiFi running NeuroEvolutionary (NE) learning and inference. We evaluate the performance of such a collaborative system and detail the compute/communication characteristics of different arrangements of the system that trade-off parallelism versus communication. 
Using insights from our analysis, we also propose algorithmic modifications to reduce communication by up to 3.6x during the learning phase to enhance scalability even further and match performance of higher end computing devices at scale.
We believe that these insights will enable algorithm-hardware co-design efforts for enabling 
continuous learning on the edge.
\end{abstract}

\begin{IEEEkeywords}
NeuroEvolution, Edge system, Learning
\end{IEEEkeywords}

\section{Introduction}
\label{sec:intro}
\insertWideFigure{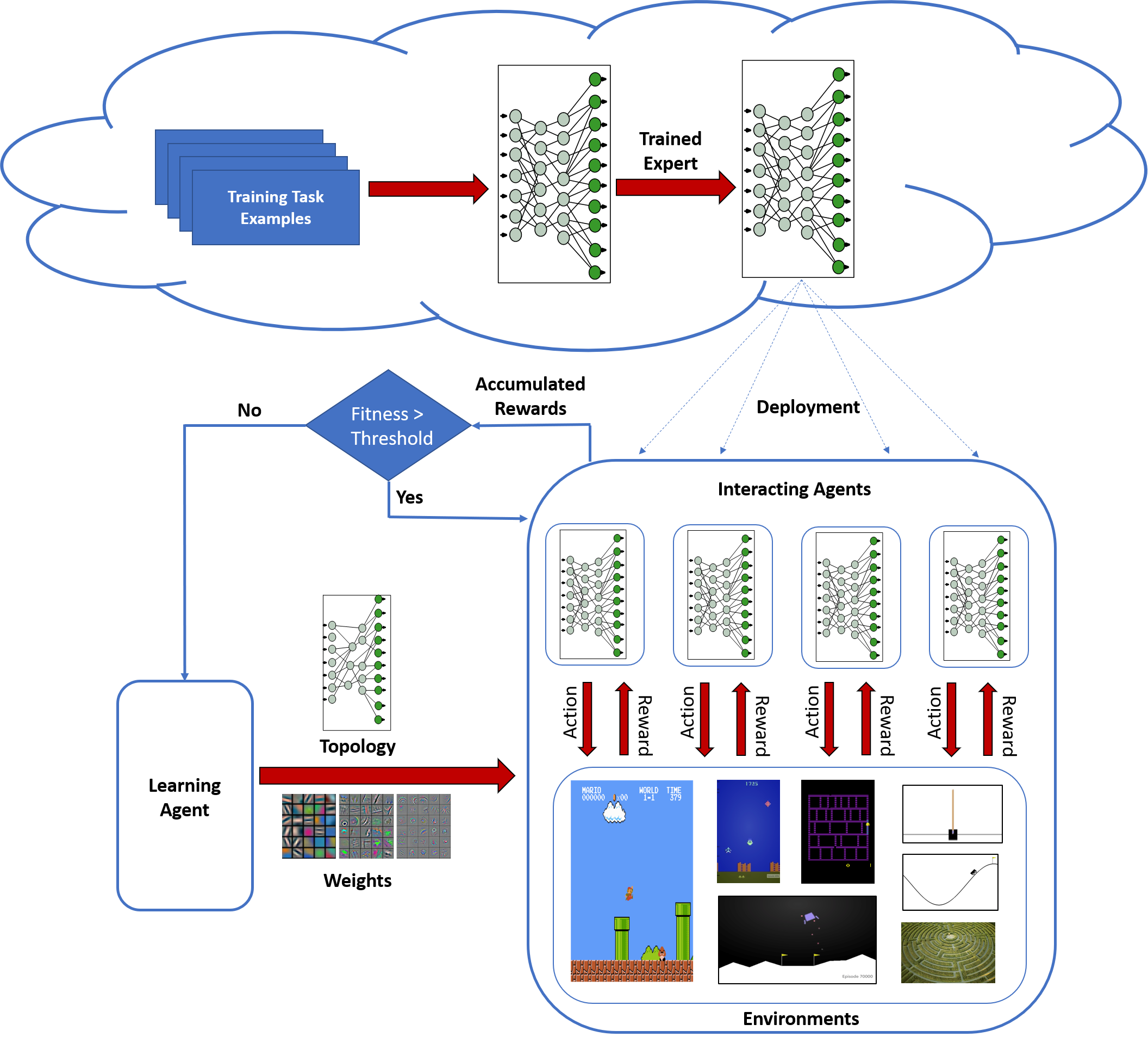}{Overview of Proposed Setup for Collaborative Learning at the Edge}

Advancements in deep learning (DL) is the defining technological milestone of the present times. Tasks like, computer vision~\cite{resnet}, speech comprehension~\cite{deepspeech} etc. 
which were deemed impossible by computers at the turn of the previous decade, are now being routinely performed at superhuman accuracy. 
These new techniques promise game-changing consequences for industry and society.

Supervised Learning, the engine powering this revolution, however has its limitations. DL solutions are only as good as its data-set and the topology it has been trained on. Prerequisites are therefore, gargantuan amounts of data, meticulous labelling and careful construction of the topology by experts. After the ingredients are in place, the model then needs to be run on a high performance computing system for training. These dependencies make DL applicable only to a limited set of problems. It is therefore quite natural that AI happens only on the cloud. The edge devices, for most cases, serve only as an interface collecting queries and displaying the results processed in the cloud. Although recent advancements in specialized accelerators have enabled inference on edge, training still largely remains confined to cloud. 

As AI becomes pervasive, there has been a growing interest in training on the edge. 
This is due to a variety of use-cases.
First - the environment of deployment can have a huge degree of variation from the data a model was trained on. An example being an agent trained to walk on the road, but encountering sand in the real world. It is impractical to expect a diverse dataset incorporating all deployment scenarios from the outset. However, transmitting all the sensed data to the cloud for re-training and choking its uplink and downlink bandwidth is not scalable as the number of edge devices grows at an exponential rate. Second - privacy~\cite{edgeai} and security concerns (especially in the context of autonomous military agents) may warrant 
little to no transmission of data to a remote server. Third - autonomous agents operating in remote regions may have intermittent or no connectivity to the cloud.
For any or all of these reasons, there is a need to develop solutions for 
enabling learning on the edge.

However, on-device learning introduces a unique set of challenges. Challenge \#1: training demands for compute and memory
far exceed those provided by embedded 
microprocessors on edge agents; 
Challenge \#2: 
it is not possible to ``label" sensed data
in order to run traditional supervised learning; Challenge \#3: as edge agents 
undertake unique tasks that they have not 
been trained for, it is not clear what DNN model they should use in the first place.



The challenges mentioned above motivate a clear need for a solution that can learn and adapt to the dynamics of new environments and changing problems on the edge. \autoref{fig:Overview} shows an example of such a system wherein a trained model/expert is deployed onto the edge and the agents adapt to the new environment/task autonomously. Each agent uses the deployed expert to perform the task at hand and continues to evaluate its fitness against a rubric as a measure of how well the expert is performing at the task. In the event of a change of task or environment, if the fitness of the expert deteriorates below a certain threshold, the agents invoke the learning process on the edge and continue to learn a new expert until the desired fitness is achieved. This new expert is used until another change occurs and learning needs to be performed again. The key question then becomes - is it possible to realize such a system today?


Federated learning~\cite{federated} attempts to solve some challenges by being able to apply small directed updates to a DNN model geared towards the unique task but are inherently limited by the topology of the predefined network, need for easily inferred labels from data along with the high compute and memory requirements of Supervised Learning approaches.

Reinforcement learning approaches have been successful in tackling Challenges 2 and 3 to some extent.
In last few years Deep Reinforcement Learning algorithms like DQN, DDQN, Duel DQN, A3C~\cite{dqn,doubledqn,dueldqn,a3c} have shown that Deep RL can learn from the environment, without a dataset and surpass humans in complex tasks like playing games involving complex strategies like DOTA~\cite{openaifive}, or develop intuitive understanding like Go~\cite{alphago}.
RL works by interacting with the environment i.e., performing an action in order to make progress towards the objective, receiving rewards, and updating its internal policy function (which is a DNN modeling the task) to improve over time.
Unfortunately, the update step requires training via backpropogation which is known to be a highly compute and memory intensive process.
This makes them incapabable of dealing with Challenge 1 directly on edge devices.


An alternative to Deep RL is a set of algorithms called Neuro-Evolutionary Algorithms (NE).
Similar to RL, these algorithms also interact with the environment and adapt to maximize the reward. However unlike Deep RL, each update step does not require back-propagation training but uses Genetic Algorithms instead to perturb the internal DNN structure and hyperparameters. 
NEs have shown promise in the cloud~\cite{salimans2017evolution, uberscale} to enhance DNN model search during training, and have demonstrated high scalability over GPUs.
Recent work~\cite{genesys} has also demonstrated custom accelerators for NE to enhance performance and energy-efficiency.

In this work, we study the viability and promise of using NE for collaborative learning over embedded devices. We quantify the computation and communication costs of these algorithms, and suggest algorithmic optimizations to enhance scalability. We demo a prototype built using Raspberry Pis that performs adaptive collaborative learning using WiFi matching the performance of higher-end computing devices at much lower energy and dollar cost. Such deployment of collaborative learning on the edge can result in a mass proliferation of autonomous robotic swarms capable of adapting to a new problem settings in a robust manner.

We make the following contributions:
\squishlist
\item{We present \textit{CLAN - Collaborative Learning using Asynchronous Neuroevolution}, 
a closed-loop collaborative learning system for the edge that runs over loosely coupled (via WiFi) 
Raspberry Pis.}
\item{This is the first work, to the best of our knowledge, demonstrating distributed neuro-evolutionary learning on real embedded hardware.}
\item{We characterize the computation, communication costs of a NE algorithm and study mechanisms to run both its inference and learning phases in a distributed manner.}
\item{We propose algorithmic modifications that allow NE to scale upto 65 nodes and show a 2 times improvement in performance over Hard Scaled NE.}
\item{Proposed modifications bring down the share of communication to 22\% vs 50\% when naively scaled as is.}
\squishend

\section{Background and Motivation}
\label{sec:background}

\subsection{Supervised Learning}
\label{subsec:sup_learning}
Conventional machine learning algorithms learn from data. 
In a machine learning flow, a model is constructed with a fixed set of parameters and initialized with some set of values.
The model is then fed with inputs from a labelled dataset to generate inference outputs which are then compared with the set of labelled correct output values and the differences are recorded. 
These differences or errors are then formulated into a loss value, which is used to update the model weight using optimization algorithms like gradient descent. 
These set of steps are called training. 

For neural networks the most widely algorithm to update the parameters is called Backpropagation(BP). 
In BP, the parameters of the neural networks are updated by repeated gradient and error calculation starting from the output layers towards the input ones, which requires activation data from the inference pass making BP extremely memory and compute intensive. 
Furthermore, the quality of training greatly depends on the quality and the size of the dataset. 
Supervised learning's dependence on training data sets and BP prohibits leveraging them 
directly to build the continuous learning edge system shown in \autoref{fig:Overview}.

\begin{table}[h]
\centering
\caption{Comparison of various learning techniques}
\footnotesize
\begin{tabular}{ | m{7.5em} | m{1.5cm} m{1.6cm} m{2.1cm} |}
\hline
            & \textbf{DL}      & \textbf{RL}      & \textbf{NE}                                                                                                                                                                                                                                                                                                                                          \\ \hline
\textbf{Data}       & \begin{tabular}[c]{@{}l@{}} Labeled \end{tabular} & \begin{tabular}[c]{@{}l@{}} Unlabeled \end{tabular} & \begin{tabular}[c]{@{}l@{}} Unlabeled \end{tabular} \\[1.5ex] \hline

\textbf{HyperParameter} \textbf{Tuning} & \begin{tabular}[c]{@{}l@{}} Manual \end{tabular} & \begin{tabular}[c]{@{}l@{}} Manual \end{tabular} & \begin{tabular}[c]{@{}l@{}} Automatic \end{tabular} \\[1.5ex] \hline

\textbf{Task Flexibility} & \begin{tabular}[c]{@{}l@{}} Task-specific \end{tabular} & \begin{tabular}[c]{@{}l@{}} Flexible reward \\ but limited to \\ architecture \end{tabular} & \begin{tabular}[c]{@{}l@{}} Flexible rewards \\ and flexible \\ architecture \end{tabular} \\[1.5ex] \hline

\textbf{Compute} & \begin{tabular}[c]{@{}l@{}} Heavy \\ BP-based \end{tabular} & \begin{tabular}[c]{@{}l@{}} Heavy \\ BP-based \end{tabular} & \begin{tabular}[c]{@{}l@{}} Low no-BP \\ Massive Parallelism \end{tabular} \\[1.5ex] \hline

\textbf{Memory} \textbf{requirements} & \begin{tabular}[c]{@{}l@{}} High \end{tabular} & \begin{tabular}[c]{@{}l@{}} High \end{tabular} & \begin{tabular}[c]{@{}l@{}} Low \end{tabular} \\[1.5ex] \hline

\end{tabular}
\label{table:ne_vs_rl}
\end{table}

\subsection{Reinforcement Learning}
\label{subsec:reinf_learning}

In layman's terms Reinforcement Learning (RL) can be depicted as learning on the fly. 
In a RL setting there is an agent which tries to learn the optimal policy to complete a task by performing repeated actions in a given environment which generates a reward value that encapsulates the effectiveness of the given action.
With each such reward obtained, the agent updates its policy such that future rewards are maximized.

In contemporary RL algorithms like DQN, A3C etc.~\cite{dqn, a3c} the policy function is approximated by a Deep Neural Network. 
This internal neural network is thus trained periodically using backpropagation, with the recorded state-pairs and their corresponding reward values as a measure of loss. The use of BP again limits the deployment of RL on the edge.


\subsection{Neuro-Evolutionary Algorithms}
\label{subsec:ea}
Neuro-evolutionary (NE) algorithms work in the same setting as RL algorithms, where the solution to a given problem is learnt by continuous interaction with the environment. 
However unlike RL there is no fixed model which is trained to learn the policy function, but the policy function is evolved using genetic algorithms. 
NE algorithms start with a population of simple neural networks. The topology and weights of these networks are then tweaked and built upon using the cross-over and mutation operations over multiple generations. 
In each generation, every member of the population (the neural networks) is given a chance to solve the given problem. 
A fitness value is then assigned based on how well they performed the given task and used to select a few of the fittest members.
These chosen members are then passed to a genetic algorithm to produce a new generation of neural networks which repeat the process.

So far NE has been confined to the cloud and is being used today, in limited form, for enhancing DNN training~\cite{uberscale, salimans2017evolution}.
However, NE may offer tremendous opportunities on 
the edge due to its high scalability~\cite{salimans2017evolution} and limited memory requirements~\cite{genesys}. 
Quantifying, analyzing and demonstrating this opportunity is the goal of this work.
\begin{table}[h]
\centering
\caption{Terminology in NEAT}
\footnotesize
\begin{tabular}{|l|l|}
\hline
\textbf{Term}       & \textbf{Meaning}                                                                                                                                                                                                                                                                                                                                             \\ \hline
\textbf{Gene}       & \begin{tabular}[c]{@{}l@{}}This is the basic building block in \\ NEAT, which can be of two types; NN\\ node (i.e. Neuron), or a connection \\ (i.e. Synapse). Each gene has an \\ associated Gene ID and appropriate \\ attributes. For example, connection\\ genes have an associated weight value,\\ an input and output node gene.\end{tabular} \\ \hline
\textbf{Genome}     & \begin{tabular}[c]{@{}l@{}}The unique collection of genes that\\ describe one NN topology is known\\ as a genome.\end{tabular}                                                                                                                                                                                                                      \\ \hline
\textbf{Population} & \begin{tabular}[c]{@{}l@{}}The group of various NN topolgies in\\ a particular generation is known as the \\ population.\end{tabular}                                                                                                                                                                                                               \\ \hline
\textbf{Generation} & \begin{tabular}[c]{@{}l@{}}One complete step in the cycle of per-\\ forming Inference and Evolution is\\ known as a generation.\end{tabular}                                                                                                                                                                                                        \\ \hline
\textbf{Species}    & \begin{tabular}[c]{@{}l@{}}Genomes with similar NN topolgies \\ are grouped under one species.\end{tabular}                                                                                                                                                                                                                                         \\ \hline
\end{tabular}
\label{table:neat_terms}
\end{table}
\begin{table*}[]
\centering
\caption{Compute Components of NEAT}
\footnotesize
\begin{tabular}{|p{0.1in}p{1.0in}|p{4in}|p{1.15in}|}
\hline
\multicolumn{1}{|l}{\textbf{}}                                                                                                           & \textbf{Inference (I)}                     & \multicolumn{2}{l|}{The step involving evaluating all genomes in the population for a task at hand.}                                                                                                                                                                                                                                                                                                                                                                                                                                                              \\ \hline
\multicolumn{1}{|c|}{\multirow{9}{*}{\begin{tabular}[c]{@{}c@{}}\\\\ \textbf{E}\\ \vspace{0.85mm} \textbf{v}\\ \vspace{0.85mm} \textbf{o}\\ \vspace{0.85mm} \textbf{l}\\ \vspace{0.85mm} \textbf{u}\\ \vspace{0.85mm} \textbf{t}\\ \vspace{0.85mm} \textbf{i}\\ \vspace{0.85mm} \textbf{o}\\ \vspace{0.85mm} \textbf{n}\end{tabular}}} & \multirow{2}{*}{\textbf {Speciation (S)}}   & \multicolumn{2}{l|}{\begin{tabular}[c]{@{}l@{}}\textit{Speciation:} The addition of new structures might not always immediately result in a better individual or increased\\ fitness as it may need time to optimize. NEAT speciates the population to protect such individuals and allow them\\ time to optimize their structure before elimination as genomes only compete within their own species. \vspace{0.8mm}\end{tabular}} \\ \cline{3-4} 
\multicolumn{1}{|c|}{}                                                                                                                   &                               & \multicolumn{2}{l|}{\begin{tabular}[c]{@{}l@{}}\textit{Fitness Sharing:} NEAT performs fitness sharing where each genome must share the fitness of their species. \\ Depending on whether this adjusted fitness is higher or lower than the population average, each species grows or \\ shrinks getting a new spawn count. \vspace{0.8mm}\end{tabular}}                                                                               \\ \cline{2-4} 
\multicolumn{1}{|c|}{}                                                                                                                   & \multirow{6}{*}{\textbf {Reproduction (R)}} & \multicolumn{2}{l|}{\begin{tabular}[c]{@{}l@{}}\textit{Crossover:} This operation picks attributes from the parent genes based on the relative fitness of the parents.\vspace{0.8mm}\end{tabular}}                                                                                                                                                                                                            \\ \cline{3-4} 
\multicolumn{1}{|c|}{}                                                                                                                   &                               & \multirow{5}{*}{\begin{tabular}[c]{@{}l@{}}\textit{Mutation:} This operation is responsible to tweaking the inherited genes. It is through \\ mutations that genomes of varying sizes and dissimilar structures can be created, \\ leading to the search of an effective topology. Mutations in NEAT are controlled by \\ predetermined probabilities and can be of a few different types as mentioned:\vspace{0.7mm}\end{tabular}}                                                               & Add Connection \vspace{0.7mm}                                                                \\ \cline{4-4} 
\multicolumn{1}{|c|}{}                                                                                                                   &                               &                                                                                                                                                                                                                                                                                                                                                                                                                                                                                  & Delete Connection  \vspace{0.7mm}                                                            \\ \cline{4-4} 
\multicolumn{1}{|c|}{}                                                                                                                   &                               &                                                                                                                                                                                                                                                                                                                                                                                                                                                                                  & Add Node  \vspace{0.7mm}                                                                     \\ \cline{4-4} 
\multicolumn{1}{|c|}{}                                                                                                                   &                               &                                                                                                                                                                                                                                                                                                                                                                                                                                                                                  & Delete Node \vspace{0.7mm}                                                                   \\ \cline{4-4} 
\multicolumn{1}{|c|}{}                                                                                                                   &                               &                                                                                                                                                                                                                                                                                                                                                                                                                                                                                  & Perturb Weights \vspace{0.7mm}                                                               \\ \cline{2-4} 
\multicolumn{1}{|c|}{}                                                                                                                   & \textbf{Generation Planning}                     & \multicolumn{2}{l|}{\begin{tabular}[c]{@{}l@{}}The step involves sorting the various members of the population according to fitness, collecting the parent pool for\\ each species, determining the number of children for each species (spawn count) and selecting parent genomes\\ for each child.\end{tabular}}                                                                                                                                                                                                                         \\ \hline
\end{tabular}
\label{table:neat_compute}
\end{table*}
\subsection{NE vs RL}
\label{subsec:ne_vs_rl}
Although NE and RL operate on a similar notion of learning from experience, similarities between the two algorithms are far fewer than the differences, especially in their compute and memory behaviors. ~\autoref{table:ne_vs_rl} summarizes some of these differences and we look at those in detail in this section and see why NE can be a viable alternative at the edge to RL.

\textbf{Memory Constraint} Edge devices are particularly constrained on memory and off-chip bandwidth and it is thus important to analyze the memory requirement of a learning algorithm to determine its viability. 

Training in Deep RL, like in DNNs require storing weight parameters and activations as an example forward propagates through the network as these must be retained to calculate the gradients during the backward propagation. Let's take a look at DQN~\cite{dqn}, a popular RL algorithm that solves the Atari environment. The model used stores about 1.7 million parameters and computes about 22 thousand activations for each run. Using 32-bit floating point storage, this amounts to a storage requirement close to 7 MB, which is already higher than the typical on-chip available memory storage. As described in the paper, using a batch size of 32 quickly increases the burden of storage to more than 220 MB.

Learning in NE however, does not need to store any activations or gradients and the storage requirements are quiet simply only the parameters of all the members of the population. Since NE networks are inherently sparse and build connections from scratch, the networks are fairly small. Previous studies~\cite{genesys} have shown the memory requirement for NEAT (a NE algorithm which we look at in detail in ~\autoref{subsubsec:neat}) is less than 1 MB including complex environments like the aforementioned Atari.

Although, the structural organization of DNNs allow more efficient encoding mechanisms than is possible for sparse networks generated by NEAT, the small memory requirement would allow even memory constrained edge devices to effectively cache all the required data.

\textbf{Compute Behavior}
Along with heavy memory requirements, BP based methods such as RL also have heavy compute requirements. 
On the other hand, NE with hundreds of small networks running every generation poses a significant question at compute capabilities as well. However, the independence of each neural network from the other offers immense parallelism opportunity which is almost absent in most RL algorithms. The sheer amount of available parallelism can be clearly seen in one of the seminal works~\cite{es} in the field where NE trained for 1 hour over 720 CPUs was able to perform the same amount of computation performed by 1-day results of A3C~\cite{a3c}, a popular RL algorithm and generate competitive solutions for Atari environments. 
Note that, NE did not beat RL at every environment but instead proved that it is a viable and a more scalable alternative to RL for some environments.

\textbf{Hyperparameter tuning and flexibility}
Building a neural network requires expert tuning of various hyperparameters including the network architecture itself, initialization, optimization schemes, learning rates and countless other algorithm-specific variables. Unlike deep RL, various types of NE algorithms can automate this process by building the architecture and learning parameters as they explore the given task. Hyperparameters to a NE algorithm can remain unchanged across different tasks and still manage to perform the task.
\\Although, both RL and NE have the advantage over DL techniques with the capabilities to work with unlabeled data and using flexible reward functions, RL is still limited to the underlying fixed DNN architecture. Whereas, NE algorithms can even modify the network architecture in response to a given task.
This is especially powerful at the edge when the exact nature of the task is not known beforehand and can change frequently as the deployment scenario differs. With NE, as long as the fitness of an individual is defined by its ability to perform the task, it can evolve at the edge to learn the task in the new environment.

It is also important to note that NE along with its advantages also has its drawbacks such as low sample efficiency, DL/RL typically have better performance in the presence of labeled datasets etc. The broad nature of tasks at the edge almost certainly means that no one algorithm is the best fit for all but looking at these various aspects, we can see that NE is certainly a viable alternative to deep RL at the edge and presents an interesting opportunity. We discuss and analyze this opportunity in the next few sections.
\insertWidePNG{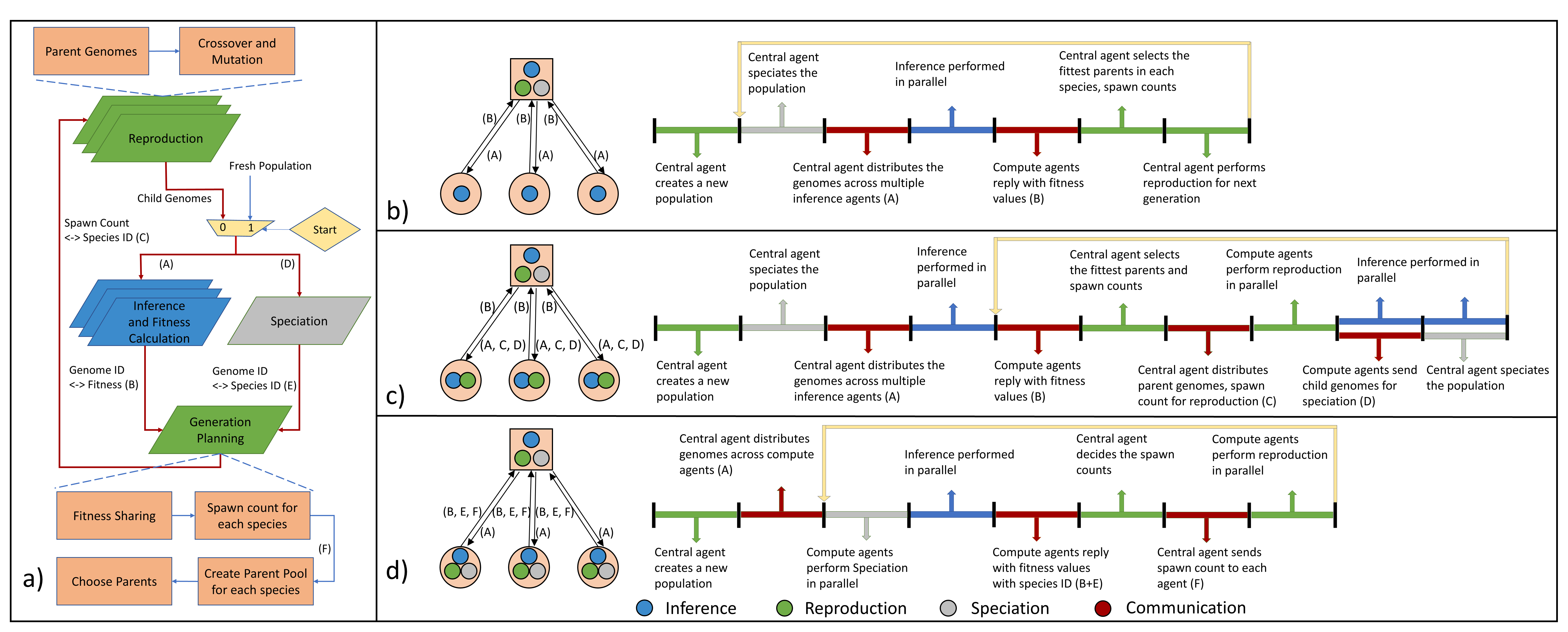}{Naming scheme of distributed system configurations in CLAN is CLAN\_$<$IRS$>$ for Inference, Reproduction and Speciation respectively where I,R can be Distributed (D) or Central (C) and S can be Synchronous(S) or Asynchronous(A). (a) Flow of NEAT Algorithm (b) CLAN\_DCS (c) CLAN\_DDS (d) CLAN\_DDA}

\section{Collaborative Learning using Asynchronous Neuroevolution (CLAN)}
\label{sec:design}
The massive parallelism available in NE and typically low compute capabilities of a single edge device call for a distributed solution to the problem. The notion that there exists a population of problem solvers in NE that each tries to learn the solution also lends itself to a distributed setting.
\subsection{Target Algorithm: NEAT}
\label{subsubsec:neat}
In this work, we leverage the NeuroEvolution of Augmented Topologies (NEAT) algorithm.~\cite{neat} that belongs to the class of Topology and Weight Evolving Artificial Neural Networks (TWEANNs). NEAT is a powerful algorithm that has been used successfully across many problems domains from learning function approximators for RL~\cite{functionapprox} to independent implementations of learning the flappy bird game \cite{flappygit}. 

We use NEAT to motivate and evaluate the scalability of NE in a distributed system setting. Through the course of this text, we will use certain terminology used in NEAT which has been mentioned in \autoref{table:neat_terms}.
\autoref{fig:design_flow}(a) shows the computation flow in NEAT as it repeats generation by generation. Different compute blocks are represented by vertices and the edges dictate communication between them. There are three major compute blocks in NEAT; namely Inference, Reproduction and Speciation, the latter two along with the process of Generation planning fall under the broad umbrella of Evolution. The different compute blocks and their roles which are critical to the discussion in this paper have been detailed in \autoref{table:neat_compute}.

GeneSys~\cite{genesys} discusses the computation and memory behavior of NEAT and finds that the memory requirements of the algorithm is small enough ($<$1 MB) to fit on on-chip memory. This is much smaller than other paradigms where training is done using BP. NEAT works by exploring the topology and weight space each generation and thus has no need to store activations/error gradients across examples during training.
Additionally, the availability of large number of members in a population trying to solve a given task independently presents massive parallelism opportunity which can be leveraged in a distributed system.
\\The low compute and memory requirements, massive available parallelism and the ability to constantly adapt to and solve complex problems makes NEAT a good candidate to deploy intelligence on the edge as well as enable distributed computing.

\insertPNG{cost_figure}{Cost analysis of (a) Inference (b) Reproduction (c) Speciation}

\insertWideFigure{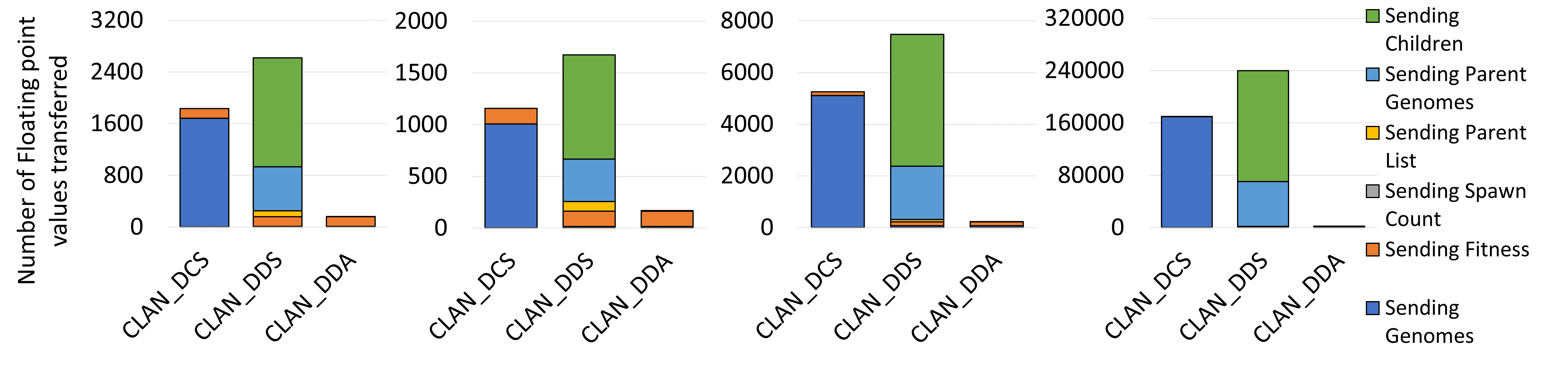}{Breakdown of Communication Cost for various configurations (a) Cartpole-v0 (b)MountainCar-v0 (c)LunarLander-v2 (d)Atari Games}

\subsection{Target Workloads}
\label{subsec:workloads}
We use a suite of workloads from OpenAI gym~\cite{openai} mentioned below to evaluate the performance of different solutions. The workloads were carefully selected to represent varying amounts of complexity; (a) Small workloads (Cartpole-v0, Mountaicar-v0), (b) Medium workloads (Lunarlander-v2) and (c) Large workloads (Atari games - Airraid-ram-v0, Amidar-ram-v0 and Alien-ram-v0). 

We use an open source implementation of NEAT~\cite{neat_python}, modifying it as needed to perform distributed computation. Each environment is limited to 200 time-steps in our experiments which can be terminated early on successful completion or failure while performing inference per generation to collect rewards. The fitness of any model is estimated using a total accumulated reward across all time-steps with minor changes for different environments.

{\bf Cost Metric.}
Genome size is naturally defined by the number of genes it contains and hence compute and communication costs grow proportionally to it, \textit{we use the number of genes processed/communicated by different compute and communication blocks as a measure of cost}.
A gene is a 32-bit datastructure~\cite{genesys} and encodes the DNN structure or hyperparameter (weights). Analyzing the costs of different compute blocks allows us to identify the areas that stand to gain the most by distributed compute. \autoref{fig:cost_figure}(a,b,c) shows the trend of various compute costs across generations for different workloads and shows that inference is the costliest operation by orders of magnitude followed by Speciation and lastly by Reproduction. This conclusion will drive our designs in the following sections.

\subsection{Convergence and Accuracy}
\label{subsec:accuracy}
OpenAI gym~\cite{openai} suite defines the convergence criteria of each environment along with the scoring metric. Any model achieving a score greater than or equal to this convergence criteria are considered equivalent and to have solved the given problem at hand sufficiently well. For example, LunarLander-v2 is considered solved at 200 points. In LunarLander-v2, moving from the top of the screen to the landing pad awards between 100-140 points and moving away from the landing pad deducts points. Landing successfully or crashing ends the episode awarding +100 and -100 points respectively. Each leg touching the ground is awarded +10 points and using the main engine adds a penalty of -0.3 points per frame. In this manner, the environment awards both accuracy of landing and the speed. Analysis of convergence and accuracy performed in ~\cite{genesys} shows us that NEAT is robust, can successfully converge a variety of workloads and maintain the accuracy over multiple generations. We now focus on understanding how each of these generations of NEAT can be computed, distributed and accelerated at the edge.

\subsection{Proposed Designs}
As we have seen in \autoref{fig:design_flow}(a), there are three major compute components - \textit{Inference, Reproduction and Speciation}.
\autoref{fig:design_flow}(b,c,d) describes three designs built in this work for collaborative learning. We assume one central, and multiple distributed devices - Raspberry Pis in our case.


\subsubsection{Hard Scaling}
%

Of the three main compute blocks, Inference and Reproduction can be performed in parallel in a distributed setting out of the box.
As we earlier discussed in \autoref{subsec:workloads}, looking at \autoref{fig:cost_figure}, it is clear that inference is orders of magnitude more compute intensive and this sort of behavior is not unexpected due to each inference step being performed over multiple time steps. Therefore, this becomes our first candidate for distribution.

\textbf{CLAN\_DCS (i.e., Distributed Inference).}
In every generation each member of the population interacts with the environment to attain its fitness score. This leads to multiple forward passes owing to multiple time steps raising the compute costs of performing inference.
However there is no dependence between the inference across genomes and thus could be performed in parallel essentially leveraging Population level parallelism (PLP).
This motivates our first design choice CLAN\_DCS, where inference for multiple genomes are performed concurrently in a distributed fashion.
Therefore, the inference step is distributed (D) while reproduction and speciation are performed centrally (C) and synchronously (S) respectively.

To achieve this, an additional step has to now be introduced which involves sending out genomes formed by reproduction to multiple compute agents for inference and subsequently gathering back the fitness values for every genome once inference steps are completed.
The configuration and time-line of compute/communication followed in such a setup has been shown in \autoref{fig:design_flow}(b).
\insertWideFigure{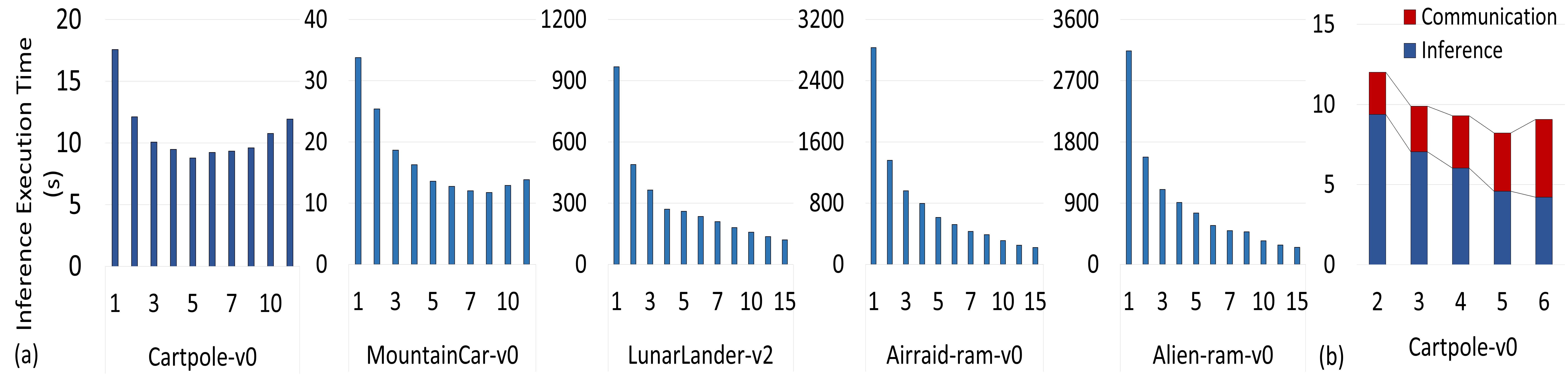}{(a) Execution Runtime at scale for Distributed Inference CLAN\_DCS (b) Breakdown of Execution Runtime between Inference Compute and Communication}
It is easy to observe that at scale, it won't take long for Amdahl's law to catch up and distributed inference, though a good start can only be as fast as the serial steps. Moreover, there does not exist a necessary condition of repeated inference over multiple time steps in the real world reducing the compute share of inference similar to the compute of evolution. The next most expensive compute block is Speciation but cannot use PLP being a synchronous operation in NEAT and this motivates us to look at the next block of compute that can leverage PLP: Reproduction.


\textbf{CLAN\_DDS (i.e., Distributed Reproduction)}
We now move on to the next candidate that can exploit parallelism - Reproduction in CLAN\_DDS where along with inference, reproduction is also distributed(DD) while speciation remains synchronous(S). We distribute the reproduction step by forming children across agents in parallel. The time-line of this implementation can be seen in \autoref{fig:design_flow}(b). We can see an additional communication block over the previous system CLAN\_DCS which is due to synchronous speciation that needs to see the genome structure of all individuals to group them into buckets. Therefore, all formed children need to be communicated from agents to a central node. The central agent also needs to communicate parent genomes to agents for reproduction (and subsequent inference) as it is not a necessary condition that the fittest genomes chosen as parents are available on any given agent. Hence, a choice attempting to naively scale reproduction involves higher communication costs due to the repeated back and forth of genomes between the agents and center. Whether this cost inhibits scaling of Evolution is a question we examine in \autoref{subsec:exp}.

\subsubsection{Soft Scaling}
We breakdown the cost of communication for CLAN\_DCS and CLAN\_DDS shown in \autoref{fig:breakdown_communication}. It is evident that despite forming and evaluating child genomes on a single node in CLAN\_DDS, the communication costs do not reduce but rather counter-intuitively increase. We see that communication of genomes including sending children for speciation and parents for reproduction represent the highest share. Reduction/removal of this cost could considerably improve the performance of the algorithm by reducing communication overhead.
\\\textbf{CLAN\_DDA (i.e., Distributed Asynchronous Speciation).}
As noted before, localized speciation is a very costly component that cannot exploit any parallelism and furthermore, it is the reason behind the elevated costs of distributing reproduction. To overcome this limitation, we propose \textit{Asynchronous Speciation (AS) aka Asynchronous NeuroEvolution} in CLAN\_DDA where inference and reproduction are distributed(DD) and speciation is performed asynchronously(A). AS refers to speciation performed on small \textit{clans} of members of the population independently instead of entire population itself. Using clans, we allow multiple agents to perform independent speciation. The configuration setup and time-line for this design choice can be seen in \autoref{fig:design_flow}(b). As can be seen, there is no communication of genomes in this design after the necessary initialization. Alternatively, we can also visualize this in \autoref{fig:breakdown_communication}, the communication cost is the least for CLAN\_DDA across the workloads.
Performing AS has to pay the cost of communicating genomes only in the first generation and then continues to pay orders of magnitude lower cost than other two design choices as shown in \autoref{fig:breakdown_communication} with CLAN\_DDS paying the highest cost.
\section{Evaluations}
\label{sec:eval}
\insertWideFigure{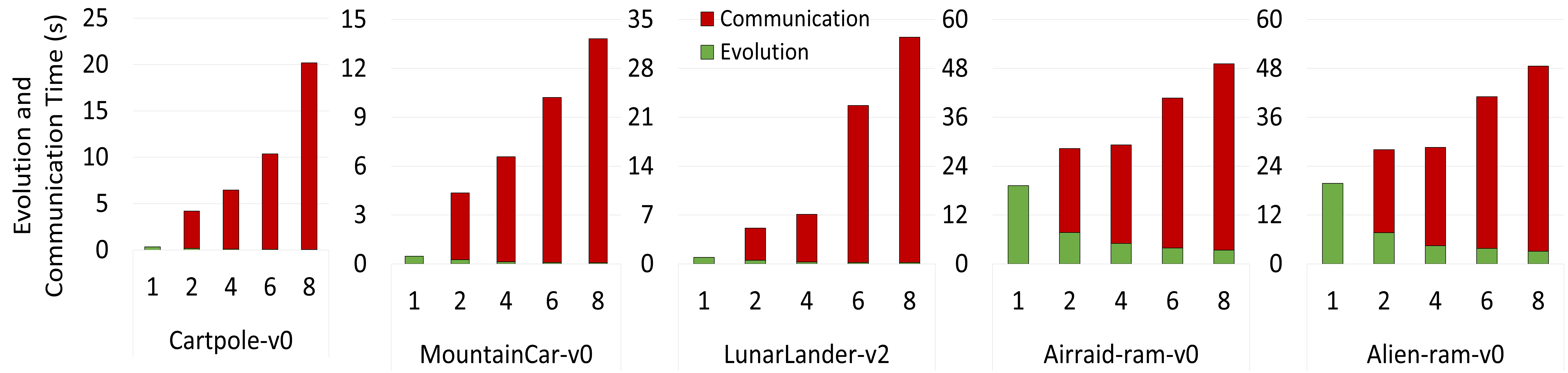}{Execution Runtime at scale for Evolution and Communication using Distributed Reproduction CLAN\_DDS}
\insertWideFigure{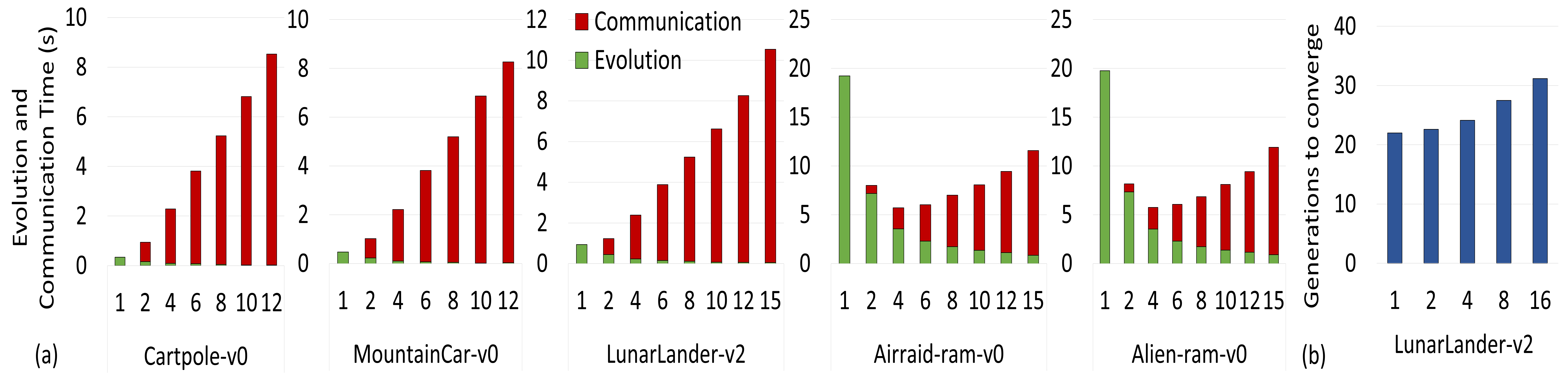}{(a) Execution Runtime at scale for Evolution and Communication using Asynchronous Speciation CLAN\_DDA (b) Accuracy vs Asynchronous nodes in CLAN\_DDA}

\subsection{Methodology}
\label{subsec:method}
Our setup is a bed of 15 RaspberryPi agents, talking over a 62.24Mbps client-to-client local WiFi network. We start from 2 active Pis and scale up gradually. The peer-to-peer latency is 8.83ms for 64B transfers.

\subsection{Comparing the three distributed settings}
\label{subsec:exp}

Note that amidar-ram-v0 results are omitted hereafter as it performs equivalently to airraid-ram-v0.
\\\textbf{CLAN\_DCS.} \autoref{fig:dist_inf} shows the inference time for Open AI workloads, in DCS setting as we scale. For small workloads (Cartpole-v0, Mountaincar-v0), scaling stops after 5 to 10 units. However, for larger workloads, the speedup is linear due to large inference run time owing to multi-timestep inference. We examine the case of single-step inference in \autoref{subsec:eval_scalability}.

On a closer inspection of compute and communication time for a smaller workload in \autoref{fig:dist_inf}(b), we notice that even though Inference continues to scale, due to similar magnitudes of inference and communication times, a small increase in agents causes the communication time to start dominating and consequently scaling stops. 
\textbf{CLAN\_DDS.} Our next set of experiments study the runtimes in CLAN\_DDS setting.
\autoref{fig:dist_rep} shows the time spent in evolution (inference runtimes are omitted for clarity as they follow the same trend as CLAN\_DCS).
Interestingly evolution does not scale beyond 2 agents, the cause of which is clear from \autoref{fig:breakdown_communication}. Communication starts to dominate from the outset since the entire population is needed to be accessed multiple times during evolution increasing the cost of communication. Even though we can see the time taken for reproduction scale accordingly, communication does not, and overpowers the scaling.

\textbf{CLAN\_DDA.} 
We see a significant benefit in runtime when communication constraints are reduced due to asynchronous speciation.
As evident for larger workloads in \autoref{fig:asyn_rep}(a), where evolution compute contributes significantly, the communication cost is not prohibitive, thus allowing evolution to scale alongside inference.

This configuration allows Evolution to scale unlike that of CLAN\_DDS however only until the overhead becomes larger than serial localized Evolution itself; a point where CLAN\_DCS could potentially prove to be a better choice. However, this is something which we have not seen till exhaustion of our test bed as execution time for Distributed Asynchronous Evolution continues to be lesser than that of localized Evolution for the larger workloads. We further this line of thought in \autoref{subsec:eval_scalability}

\insertWideFigure{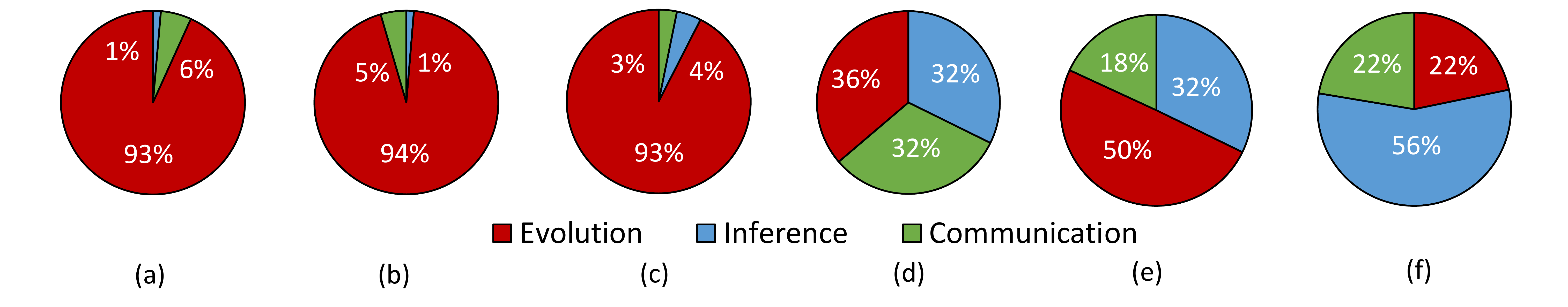}{Breakdown of compute share in different designs for single-step inference with two nodes for Cartpole-v0 (a)CLAN\_DCS (b)CLAN\_DDS (c)CLAN\_DDA and Airraid-ram-v0 (d)CLAN\_DCS (e)CLAN\_DDS (f)CLAN\_DDA}
\subsection{Impact on Accuracy with Asynchronous Speciation}
\label{subsec:acc_spec}
It is important to evaluate the the effect of performing speciation on \textit{Clans} instead of the entire population. Performing Asynchronous Speciation can potentially lower the rate of exploration by reducing competition. A comprehensive study of the effect of AS on the accuracy and convergence time of various environments is an interesting question that needs further inspection.

To evaluate this, we consider the LunarLander-v2 workload and evaluate the convergence accuracy with an increasing number of Clans and a population size of 150 members. A single Clan represents Synchronous Speciation, as described in~\cite{neat}. We perform 10 runs and average the convergence accuracy achieved at each data point to handle the probabilistic nature of the algorithm.

We can see in \autoref{fig:asyn_rep}(b), the effect of increasing number of Clans and as expected, the number of generations needed to converge gradually increases. This shows that at scale, the convergence accuracy drops, albeit slowly when using AS.
The presence of such a trade-off space between accuracy and compute performance is an interesting area to explore. One can think of many ways to mitigate this problem such as allowing periodic global speciation, and is an idea ripe for future work.





\subsection{Evaluating Scalability}
\label{subsec:eval_scalability}
In our evaluations so far, we have operated all workloads with inference lasting multiple time steps. However, this assumption does not always hold true outside of typical RL game workloads such as using NE in autonomous robotics. So far, because of this nature, inference compute has dominated heavily and not allowed our evaluations a chance to understand the difference in performance of the complete learning process under different configurations. 
We overcome this limitation by evaluating each genome only once in a given generation, thus limiting the dominance of inference. In such a scenario, we can truly test the mettle of various configurations.
\insertFigure{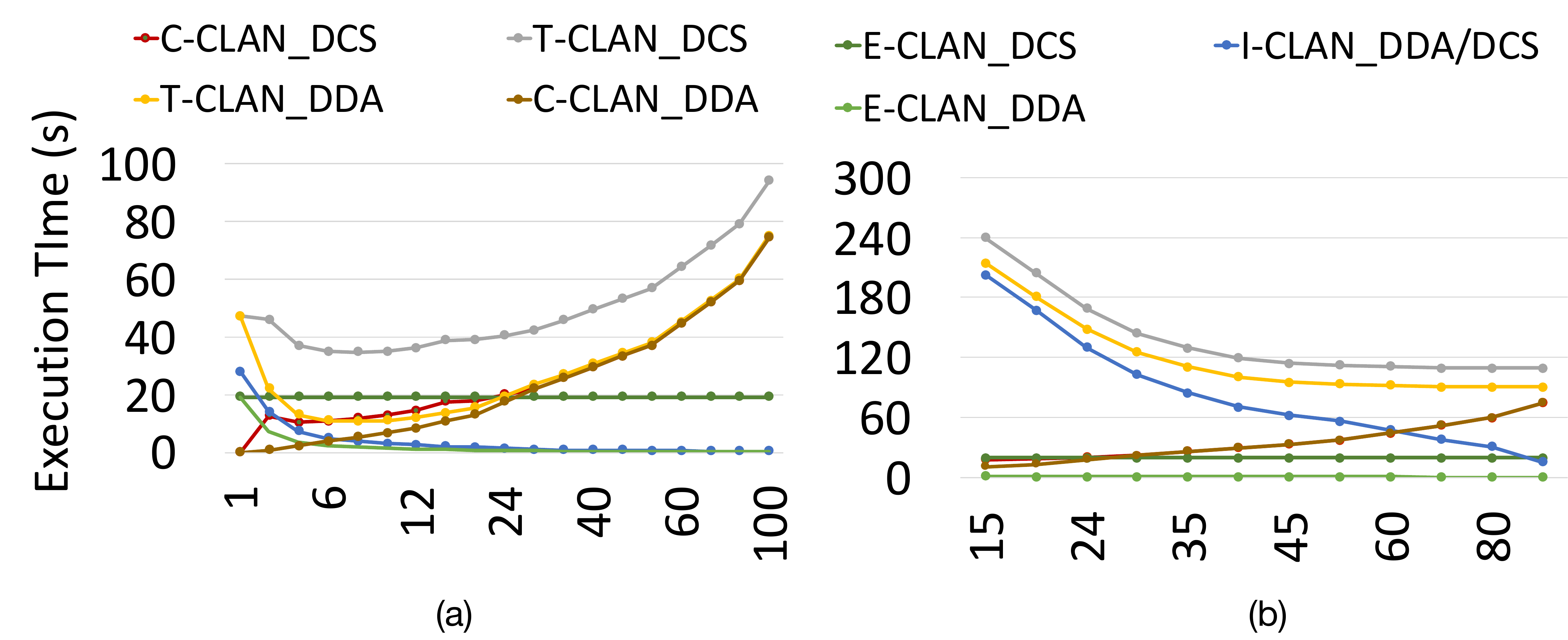}{Extrapolation study up to 100 units for Airraid-ram-v0 for (a)Single-step Inference (b)Multi-step Inference. C=Communication, E=Evolution, T=Total}
\insertFigure{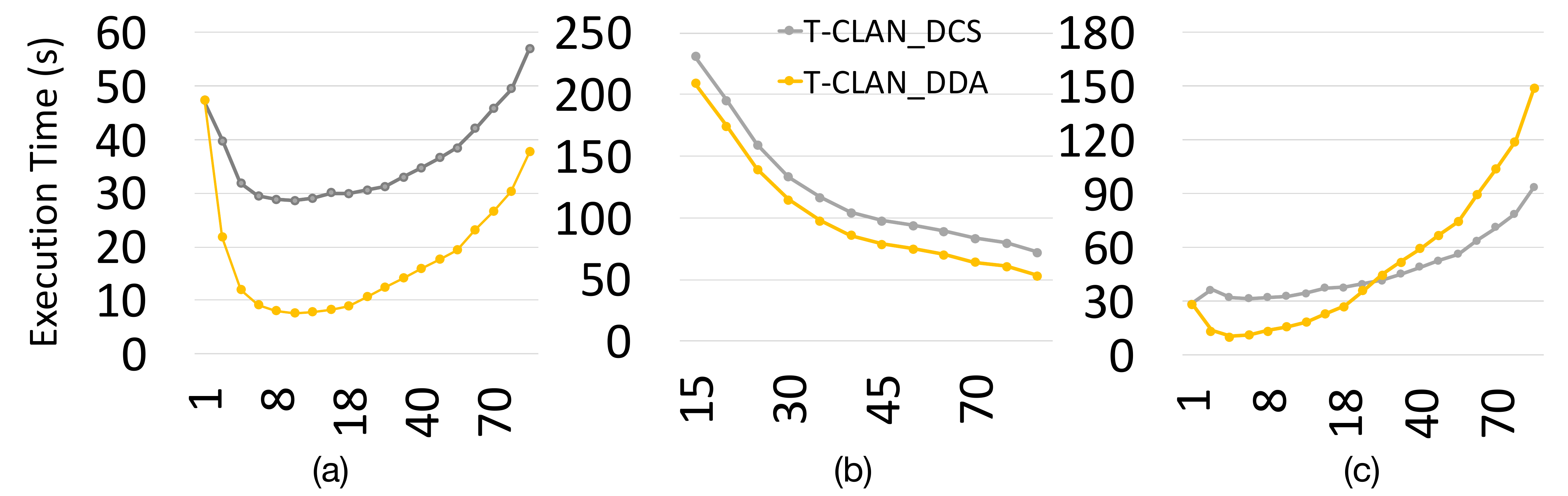}{Impact on scalability due to changing technology trends for Airraid-ram-v0 in (a) Communication on single-step inference (b) Communication on multi-step inference and (c) Hardware platform on multi-step inference}
\insertWideFigure{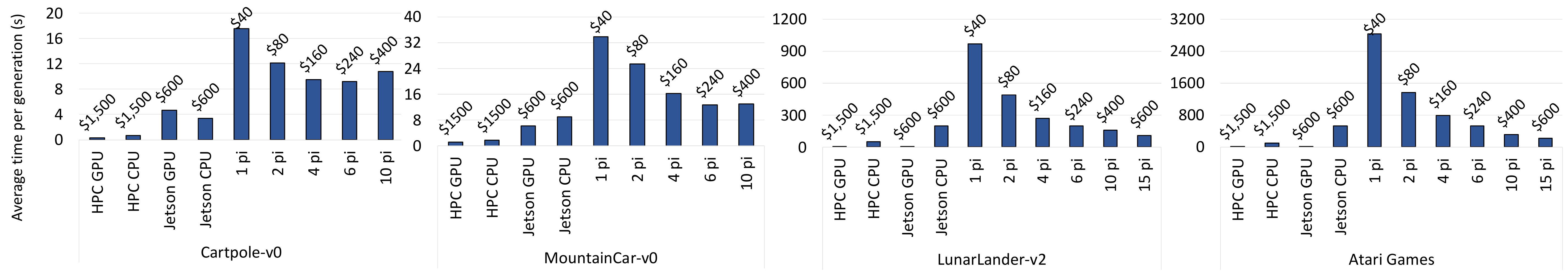}{Comparing Platforms - Performance per Dollar}
We use two workloads; one from each class of large and small workloads namely Airraid-ram-v0 and Cartpole-v0 respectively.
The share of compute/communication using two nodes for both workloads has been shown in \autoref{fig:designcomparison} for comparison and the difference will only become more prominent as node count increases.
The effect of the increased amount of communication in CLAN\_DDS becomes more evident here. Along with the increased amount of communication involved, the constant cost of invoking the communication channels also kills this design when such a scenario is presented where the amount of compute is no longer sufficient to amortize the constant setup costs. Looking at \autoref{fig:designcomparison}, it can be seen that communication consumes about 50\% and 94\% of the share for larger and smaller workload respectively in the CLAN\_DDS configuration. This share reduces to 36\% and less significantly to 93\% respectively in CLAN\_DCS. The best result is seen while using CLAN\_DDA,  where the share of communication is only 22\% (3.6 times lesser than CLAN\_DDS) and 93\% respectively indicating both significant energy and runtime savings.
This result is significantly important to note as the cost of communication can get extremely high between edge devices using slower and more distant mediums of communication or go the other way by using better and faster mediums, an effect which we discuss in \autoref{subsec:extrapolation}.

\textbf{Indefinite Scaling}
An interesting question in evaluating scalability is how far can we push before adding nodes does not add performance, or worse when a serial implementation becomes a better choice.
So far, we have been restricted by our test-bed of 15 units.
We attempt to go beyond this restriction 
by extrapolating trends.


For larger workloads such as those of Atari games, we did not notice a point where scalability stops for Inference. Whereas for Evolution, despite noting an inflection point while using CLAN\_DDA, we do not yet see a point where a serial choice would be better under the limitation of our test-bed.

We wish to assess at what point does scaling stop for different configurations and which configuration proves to be a better choice. We evaluate both multi-step and single-step inference as both operate under two different interesting situations.
We use trends of inference, evolution and communication overheads in each configuration to extrapolate these curves. For more accurate extrapolation of trends, we performed experiments using reduced population sizes, effectively emulating higher scale as each agent performs computation (and communicates) for a fewer genomes.
We do not study CLAN\_DDS as it is clear that it performs worse in any setting. The extrapolated curves for both multi-step and single-step inference have been shown in \autoref{fig:extrapol_1}. We show raw inference, evolution compute along with communication overheads and also plot the total execution time.
The key observation here is that for both multi and single-step inference, CLAN\_DDA always performs better than CLAN\_DCS in total time taken. It is also worthwhile to note that as raw compute continues to scale due to the available PLP, the total time taken is defined by communication overhead. For multi-step inference, the performance stagnates around 50 units for both configurations with CLAN\_DDA performing better by 1.1x throughout the scale. For single-step inference, the improvement in performance for both configurations stops at 10 units. However, CLAN\_DCS becomes worse than a serial implementation at 40 units whereas CLAN\_DDA can push this limit to 65 units performing 2x better on average across the scale.

\subsection{Impact of Technology and Hardware}
\label{subsec:extrapolation}
There have been two key points of discussion throughout, i.e. Compute and Communication. Next, we discuss the impact of changes in both these paradigms on scalability.


\textbf{What if the communication technology used was better?} 
From the growing interest on ML for IoT~\cite{mohammadi2018deep} to autonomous driving using V2V communication~\cite{al2014comprehensive}, the discussion of communication between devices has been prevalent in both academia and industry and innovations made could certainly be leveraged in the future in a setting like ours. To analyze the effect, we halve the communication cost as an approximation and plot the total execution time curves for CLAN\_DCS and CLAN\_DDA for both single-step and multi-step inference in \autoref{fig:extrapol_2}(a) and (b) respectively.
We notice the scalability of both configurations improves from 10 to 12 nodes in single-step inference. In multi-step, reduction in communication overhead allows scaling to continue through the scale without stagnation.

\textbf{What if we used Custom HW instead of Raspberry Pi?} Needless to say, there has been a tremendous surge of research work in the field of custom DNN inference hardware in recent years. Many DNN accelerators have been proposed ~\cite{tpu, chen2014diannao, dadiannao, shidiannao, zhang2015optimizing, eyeriss, parashar2017scnn, eie, maeri}. Going by the trend, it is not far fetched to imagine the availability of these accelerators as commodity embedded hardware. We assume a 32x32 systolic array implementation and evaluate performance using SCALE-sim~\cite{scalesim} and plot the total runtimes for multi-step inference for both configurations under this assumption in \autoref{fig:extrapol_2}(c).
\\Signicantly faster compute performance means communication becomes a more serious issue. CLAN\_DCS cannot scale under such a situation. CLAN\_DDA however, still shows scalability scaling upto 7 nodes showing a performance improvement of over 2.5x in comparison. 
However, it is also interesting to see that CLAN\_DCS proves to be a better choice at 30 nodes where the performance of CLAN\_DDA has also become worse than a serial implementation.

\subsection{Performance per dollar}
\label{perf_dollar}

Deploying massive intelligence on the edge needs an additional metric evaluated, i.e. the price for performance. Therefore, we also compare the results of CLAN to two localized implementations, (a) High-Performance Machine (HPC) and (b) Nvidia Jetson Tx2 described in \autoref{table:cost_table}. The price of HPC machine and Jetson is comparable to 40x and 15x to the cost of a RPi respectively. We examine whether such a distributed system at scale can achieve similar performances to these much more expensive platforms in \autoref{fig:platform_comparison} while paying the communication latency and if yes, at what scale.
Performance is not comparable for extremely small workloads such as \textit{Cartpole-v0} as the communication overhead cannot be amortized by low amount of compute. However, for larger workloads, we see interesting results. At a scale of 6 compute units, the system achieves performance similar to the Jetson board, a \textit{Price-Performance Product (PPP)} improvement of 2.5x. Further scaling to 15 units, we can compare with the HPC system, achieving PPP benefit of 1.2x. The GPU performances of both the higher end platforms could not be rivaled within the limits of our single core experiments with our test-bed.
\begin{table}[h]
	\centering
	\footnotesize
	\caption{Platform Specifications}
	\begin{tabular}{lll}
	\hline
		Platform & Processor & Price\\
	\hline
		HPC CPU & 6th gen i7 & \$1500 \\
		HPC GPU & Nvidia GTX 1080 & \$1500 \\
		Jetson Tx2 CPU & ARM Cortex A57 & \$600 \\
		Jetson Tx2 GPU & Pascal & \$600 \\
        Raspberry Pi CPU & ARM Cortex A53 & \$40 \\
	\hline
	\end{tabular}
	\label{table:cost_table}
\end{table}

\section{Related Work}
\label{sec:related}
\textbf{Distributed DNN Computation}
Apart from related works on distributed training of DNN ~\cite{largednn, zinkevich2010parallelized, distributedtrain, hogwild, su2015experiments, firecaffe, strom2015scalable}, there has been significant research in distributing the inference of DNNs across devices. Significant work done~\cite{chung2018serving, hazelwood2018applied} has been on high performance computing clusters, typically using many CPU-GPU heterogenous nodes whereas we aim to distribute computation on the edge. Since edge devices typically face very different compute and energy constraints, and do not have high-speed interconnects, the trade-off space is naturally dissimilar. 
Some works such as~\cite{ddnn, hybrid, neurosurgeon} focus on distributing computation between an edge device(s) and the cloud and are thus dependent on the availability of a cloud service. Along with accelerating inference on edge devices~\cite{2018efficient, xu2018scaling} there has been some work done in distributing DNNs on the edge devices~\cite{modnn, jiashen} but such techniques are dependent on DNN primitives such as convolution layers and cannot be extended directly to the computation graph of NeuroEvolutionary algorithms.

\textbf{Distribution of EAs} 
Distributing EAs is an exciting challenge and has been explored by researchers in ORNL, Uber AI~\cite{ornl, uberscale} but on HPC systems spanning up to hundreds of CPUs. As mentioned before, HPC systems do not operate under the same constraints as an edge device. However, we believe that some insights from our work such as using Asynchronous Speciation can be leveraged even in HPC environments and hence has a broader scope.

\textbf{Custom HW platforms}
Accelerators for DNN inference have been an increasing trend in the research community~\cite{tpu, chen2014diannao, dadiannao, shidiannao, zhang2015optimizing, eyeriss, parashar2017scnn, eie, maeri, zhang2015optimizing, whatmough201714} and ASICs such as GeneSys~\cite{genesys} have demonstrated that there is tremendous parallelism available in EAs and can be leveraged to speed-up computation efficiently. This work leverages that insight to distribute computation across many devices and on the other hand, some insights from our work are agnostic to the implementation platform and could even leverage custom HW platforms as a node in the system speeding up computation significantly and allowing researchers to target even more complex problems with EAs.
\section{Conclusion}
\label{sec:conclusion}

Bringing true intelligence that is adaptive and robust to the edge on commodity available hardware can rapidly change the dynamic and the way we experience AI today. This work makes a promising contribution towards this direction by demonstrating a system of agents running on Raspberry Pis learning collaboratively using neuro-evolutionary algorithms.
To the best of our knowledge this work is the first effort to look at developing intelligence on edge using commodity devices. 

We leverage the parallelism offered by the algorithm in a distributed setting to enable complex workloads in the timing and energy constraints of edge use cases. 
We explore the challenges associated with this and propose methodologies for efficient scaling.
Proposed modifications to the algorithm allow scaling to continue up to 65 Raspberry Pi nodes showing a 2x performance improvement over naive scaling techniques while reducing communication by over 3.6 times. The proposed system using cheap Raspberry Pi hardware can outperform higher end computing platforms showcasing a Price-Performance Product benefit of 2.5x.
\section*{Acknowledgment}

This work was supported by NSF CRII Grant 1755876.
We would like to thank Fei Wu for sharing his experience on setting up embedded platforms in a distributed fashion. Finally, we would like to thank the anonymous reviewers for their insightful comments and suggestions.
\newpage
\bibliographystyle{ieeetr}
\bibliography{references}

\end{document}